\pdfoutput=1

\documentclass[11pt]{article}

\usepackage[]{emnlp2021}

\usepackage{times}
\usepackage{latexsym}
\usepackage{color, colortbl}
\usepackage[T1]{fontenc}

\usepackage[utf8]{inputenc}

\usepackage{multirow}
\usepackage{amsmath}
\usepackage{amssymb}
\usepackage{graphicx}

\newcommand{\dialog}{\mathcal{D}} 
\newcommand{\instr}{\mathcal{I}} 
\newcommand{\kb}{\mathcal{KB}} 
\newcommand{\context}{c} 
\newcommand{\user}{u} 
\newcommand{\userc}{\bar{u}} 
\newcommand{\agent}{a} 
\newcommand{\agentc}{\bar{a}} 
\newcommand{\score}{s}
\newcommand{\scoreu}{s^u} 
\newcommand{\scorea}{s^a} 
\newcommand{\tokenq}{w^q} 

\newcommand{\poolu}{P^u}
\newcommand{\poola}{P^a}
\newcommand{\query}{q}

\newcommand{\retrieved}{e}

\definecolor{darkgreen}{rgb}{0, 0.25, 0.}
\definecolor{darkblue}{rgb}{0, 0., 0.35}
\definecolor{LightCyan}{rgb}{0.88,1,1}

\usepackage{microtype}

\title{Simulated Chats for Building Dialog Systems: Learning to Generate Conversations from Instructions}

\author {
        Biswesh Mohapatra \textsuperscript{\rm 1,2,}\thanks{\hspace{0.1cm} Work done during internship at IBM Research AI},
        Gaurav Pandey \textsuperscript{\rm 1},
        Danish Contractor \textsuperscript{\rm 1},
        Sachindra Joshi \textsuperscript{\rm 1} \\
    \textsuperscript{\rm 1} IBM Research AI, New Delhi \\
    \textsuperscript{\rm 2} International Institute of Information Technology, Bangalore \\
    biswesh.mohapatra@iiitb.org \\
    $\{$gpandey1, dcontrac, jsachind$\}$@in.ibm.com
}

\begin{document}
\maketitle
\begin{abstract}
Popular dialog data sets such as MultiWOZ \cite{Multiwoz} are created by providing crowd workers an {\em instruction}, expressed in natural language, that describes the task to be accomplished. Crowd workers play the role of a {\em user} and an {\em agent} to generate dialogs to accomplish tasks involving booking restaurant tables, 
calling a taxi etc. 
In this paper, we present a data creation strategy that uses the pre-trained language model, GPT2 \cite{GPT2}, to {\em simulate} the interaction between crowd workers by creating a {\em user} bot and an {\em agent} bot.  We train the simulators using a smaller percentage of actual crowd-generated conversations and their corresponding instructions. We demonstrate that by using the simulated data, we achieve significant improvements in low-resource settings on two publicly available datasets - MultiWOZ dataset \cite{Multiwoz} and the Persona chat dataset \cite{Persona}. 
\end{abstract}

\begin{figure}[ht!]
\scriptsize
\noindent\fbox{
    \parbox{.975\linewidth}{
        \textbf{Instruction} :
        You are looking for a \textbf{train}.
        The train should \textbf{arrive by 11:45} and should leave on \textbf{Saturday}. 
        The train should go to \textbf{Cambridge} and should depart from  \textbf{Ely} . 
        Once you find the train you want to make a booking for \textbf{8 people}. 
        Make sure you get the \textbf{reference number}. 
    }
}
\noindent\fbox{%
    \parbox{.99\linewidth}{%
        CHAT GENERATED BY CROWD-WORKERS: \\ 
        \textbf{User:} \textcolor{darkgreen}{I need to find a train for \textbf{Saturday}}. \\
        \textbf{Agent:} \textcolor{darkblue}{I have many trains that depart \textbf{saturday}. Where will you be departing from and where is your destination?}\\
        \textbf{User:}  \textcolor{darkgreen}{I will be departing from \textbf{Ely} and traveling to \textbf{Cambridge}.} \\
        \textbf{Agent:}  \textcolor{darkblue}{Okay, and what time do you want to leave after or arrive by?}\\
        \textbf{User:}  \textcolor{darkgreen}{I need to arrive by \textbf{11:45}.} \\
        \textbf{Agent:}  \textcolor{darkblue}{There are 3 trains here. Do you want me to book any tickets?} \\
        \textbf{User:}  \textcolor{darkgreen}{Yes please. I need it to be booked for \textbf{8} people.} \\
        \textbf{Agent:}  \textcolor{darkblue}{Booking was successful, the total fee is 28.16 GBP payable at the station .Reference number is : ZVERHBT3.} \\
        \textbf{User:}  \textcolor{darkgreen}{Thank you, that is all I need.} \\
        \textbf{Agent:}  \textcolor{darkblue}{You're welcome, thank you for calling the Cambridge TownInfo centre. Have a great day.}
    }%
}
\noindent\fbox{%
    \parbox{0.99\linewidth}{%
        CHAT GENERATED BY SIMULATORS: \\
        \textbf{User:} \textcolor{darkgreen}{I am looking for a train to \textbf{Cambridge}.} \\
        \textbf{Agent:} \textcolor{darkblue}{There are many trains available. Where will you be departing from?}\\
        \textbf{User:} \textcolor{darkgreen}{I will be departing from \textbf{Ely}.} \\
        \textbf{Agent:} \textcolor{darkblue}{How about TR7994 that leaves at \textbf{05:35}?}\\
        \textbf{User:} \textcolor{darkgreen}{That would work, can you book it for \textbf{8} people?} \\
        \textbf{Agent:} \textcolor{darkblue}{I have booked \textbf{8} tickets on that train, the total fee is 35.2 gbp payable at the station. Reference number is ZXQSGIN7.} \\
        \textbf{User:} \textcolor{darkgreen}{thank you. that is all i need.} \\
        \textbf{Agent:} \textcolor{darkblue}{You are quite welcome. I am glad I could help. Enjoy your trip !}
    }%
}
\caption{An instruction from the MultiWOZ dialog corpus~\cite{Multiwoz} and the corresponding conversation generated by real crowd-sourced workers and crowd-worker simulators.}
\label{fig:examples}
\end{figure}

\section{Introduction}

Building dialog systems typically requires a large collection of conversation logs for training models. There exists a scarcity of the data required to train a dialog system for most tasks. Various methods have been proposed to tackle this issue including paraphrase techniques to generate artificial training data \cite{kumar2021data, syntax-change}, generating annotations including intent-slots and dialog acts \cite{slot-augment2,slot-augment, dialog-act-augmentation}
or even injecting noise to improve robustness in dialog act prediction for ASR data \cite{noise-injection}. 

\begin{figure*}[t]
    \centering
    \includegraphics[scale=0.43]{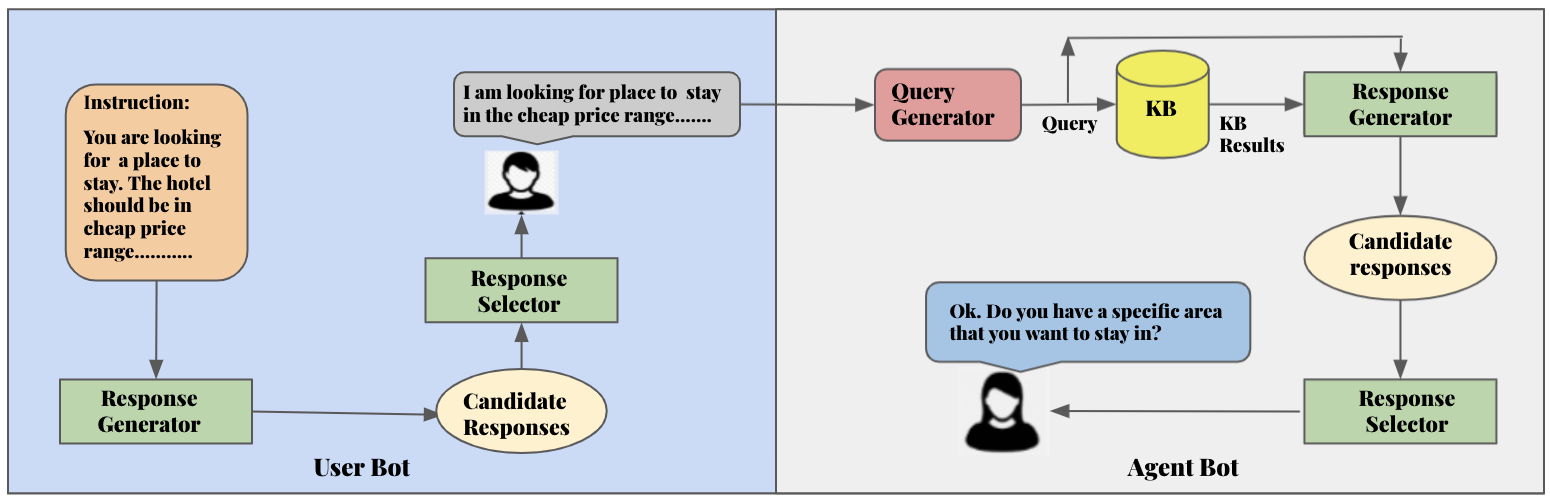}
    \caption{Generation of a conversation conditioned on the instructions and knowledge base (KB).
    Although not displayed in this diagram, each module (in green) also receives the dialog history as input. 
    }
    \label{fig:user_and_agent_bots}
\end{figure*}

Crowd-sourcing is a popular method for generating new large datasets. 
For instance, to create datasets for task oriented dialogs, crowd workers may be provided with  {\em instructions} that describes the task; workers then play the roles of a {\em user} and an {\em agent} to generate conversations \cite{Multiwoz}. The {\em user} worker begins the conversation by stating their requirement and the {\em agent} worker provides information to the user by querying a knowledge base (KB), if required. Together, both workers interact with each other via natural language to generate conversations. Similarly in Persona chat \cite{Persona}, the workers are provided different {\em personalities} to role play conversations. However, creating large crowd-sourced datasets can be time consuming and expensive.

Pre-trained transformer-based language models such as GPT-2~\cite{GPT2}, that are trained on a large number of documents crawled from the web have achieved extensive generalization in natural language understanding and generation across a variety of diverse tasks \cite{Multiwoz,rajpurkar-etal-2016-squad, dnli}. Recent works have exploited the prior knowledge in these models to train effective models for machine translation~\cite{araabi-monz-2020-optimizing}, language understanding in low resource settings~\cite{dou-etal-2019-investigating} and few-shot language models~\cite{few-shot}. 

In this paper we demonstrate how such large pre-trained models can also be used to follow {\em instructions} and generate conversations. We create a {\em user} simulator and an {\em agent} simulator. The {\em user} simulator has access to the {\em instructions} while the {\em agent} simulator has access to a knowledge base (KB). The {\em agent} simulator maps the current dialog context to a {\em belief state} (query), that can be executed over a knowledge base (KB), to retrieve a set of results if required. Thus, the simulators are trained to interact with each other to generate conversations conditioned on the instructions and the KB. In our work we train these simulators using just 5-20\% of crowd-sourced conversations by fine-tuning the pre-trained language models --- GPT2~\cite{GPT2} and Longformer~\cite{beltagy2020longformer}. We use the external knowledge present in these language models to help generate effective artificial data on low-resourced datasets. An example of a generated conversation is shown in Figure~\ref{fig:examples}.

Our experiments further show that from a small number of existing conversations we are able to train meaningful {\em user} and {\em agent} bots that in-turn generate new conversations. This in principle, is somewhat similar to a noisy student-teacher model~\cite{noisy-student} where a weaker teacher model is used to generate labels(dialogs in our case) which is then used to train a new student model that significantly outperforms the teacher model in end task. Due to its simplicity and generality, our model could be used on a wide variety of dialog systems by taking different forms of instructions.  

\noindent {\bf Contributions: } (1) We present a novel technique that effectively uses weak generative models to create new artificial data which are used to train final end task models (2) We introduce a simple yet effective dialog-generation framework\footnote{http://ibm.biz/simulatedchats} that mimics the roles played by crowd workers to generate complete conversations. (3) We demonstrate the generality of our model by generating data for two different types of dialog tasks - task oriented conversations and persona-guided conversations. We show that pre-trained language models can be successfully used for generating artificial data in low resource dialog settings leading to a 7-13\% improvement in combined score in MultiWOZ 2.0  and 2-10\% improvement in Hits@1 metric in  Persona Chat. (4) We present a human-study to assess the quality of our simulated dialogs. We find that the generated conversations are grammatically sound and meaningfully move the conversations forward.  

\begin{figure*}[ht]
    \centering
    \includegraphics[width=\textwidth,height=55mm]{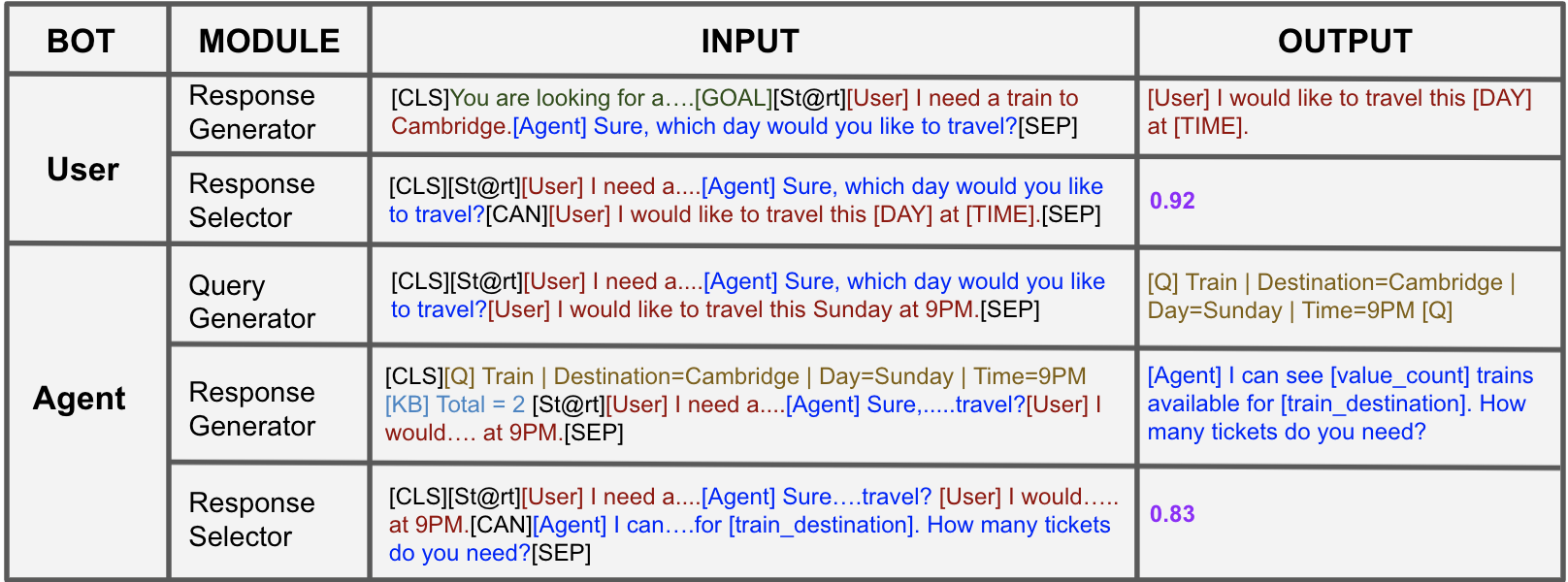}
    \caption{The input and output formats for User and Agent Bot. [GOAL] indicates end of instruction, [CAN] indicates candidate for selector, [St@rt] indicates start of conversation, [Q] and [KB] indicate Queries and KBs.}
    \label{fig:input_output_format}
\end{figure*}

\section{Related Work}
\label{sec:related-work}
The method of interacting different models to converse with each other has seen some recent successes \cite{m2m, collaborative}. \cite{corpus-free} has previously used simulators to generate conversational artificial data. However the work uses hand-crafted templates for generating dialogs. Our approach is more general and less cumbersome as demonstrated through the effectiveness of our approach on two different tasks. \cite{user-in-loop} tries to use a similar simulator approach but uses human in the loop in order to bring variations to the dialogs. On the other hand, our approach doesn't require any human involvement apart from providing diverse instructions which are easy to produce in large quantities. 

Unlike existing data augmentation methods, like those based on paraphrase generation \cite{vae-augment, PARG, lambada}, simulators create completely new conversations which create more diverse examples, helping train better end-task models (Section \ref{experiments}). 

Teacher model has been used to train student models, based on the idea of knowledge distillation \cite{Hinton2015DistillingTK}, teach dynamic loss functions \cite{teacher-loss} or for adaptation under meta-learning settings \cite{Qian_Wei_Yu_2021}. To the best of our knowledge, we are the first to adopt such a model for data generation in dialog systems. 

\section{Model}
\label{sec:model}
In order to generate the data, we train our GPT-2 based agent and user bots(teacher models) using a subset of original data to simulate low-resource environment(5\% or 20\%). In case of task oriented dialogs, we require a third model for generating belief states as well. Longformer based selector models are trained to chose from a list of responses generated by the teacher generator models as seen in Figure~\ref{fig:user_and_agent_bots}. Note that all the modules in figure (shown in green) also receive dialog history as input which has not been shown in the figure for ease of presentation. Finally the generated data from the process is mixed with the original low resourced data to create a new dataset. To test the effectiveness of new dataset, we compare the performance of newly trained student model on the new data to that of the teacher model and other baseline models on the respective end tasks of the datasets.

\subsection{Overview}
We assume that the dialog comprises of a sequence of utterances between a user and an agent i.e. $\dialog=(\user_1, \agent_1, \ldots, \user_n, \agent_n)$ where $\user_i$ is a user utterance while $\agent_i$ is an agent utterance. A turn is a pair of user and agent utterance. At any given turn $m$, the sequence of utterances prior to the turn, that is, $\context_m=(\user_1, \agent_1, \ldots, \user_{m-1}, \agent_{m-1})$ is referred to as dialog context or dialog history. Apart from the dialog $\dialog$, we have access to a set of {\em instructions} $\instr$ and a knowledge base $\kb$. The aim is to learn a model that can generate the dialog $\dialog$ conditioned on the instructions $\instr$ and the knowledge base $\kb$. That is, we wish to model $p(\dialog|\instr, \kb)$.

The dialog generation framework mimics the human-to-human data collection approach used in MultiWOZ~\cite{Multiwoz}. The dialog is generated in a sequence of turns. The user bot has access to instructions $\instr$ while the agent bot can query the knowledge base $\kb$. Thus, the joint distribution of the dialog decomposes as follows:
\begin{equation}
    p(\dialog|\instr, \kb) = \prod_{i=1}^n p(\user_i|\context_i, \instr) p(\agent_i|\context_i, \user_i, \kb) \,.
\end{equation}
The dialog history for the first turn, $\context_1$, is an empty set. The first factor in the product on the left corresponds to user bot which conditions on the instructions, as well as, the dialog history to output the user utterance. The second factor models the distribution of the agent bot over the responses, conditioned on the dialog history and knowledge base. A pictorial representation of the interaction between the two bots is shown in Figure~\ref{fig:user_and_agent_bots}. We discuss the various modules of both the bots in further detail below. The input and output formats for the various networks of these modules are shown in Figure~\ref{fig:input_output_format}.

\subsection{User Bot}
The user bot generates utterances conditioned on the dialog history and the instructions, that is, it models $p(\user_i|\context_i, \instr)$. For the sake of readability, we will remove the turn index $i$ from the distribution.
As shown in Figure~\ref{fig:user_and_agent_bots}, this distribution is modeled in two steps. Firstly, the dialog history and the instructions are fed to a {\em response generator} module which outputs a pool of candidate responses $\poolu=(\userc_1, \ldots, \userc_r)$. A {\em response selector} module then assigns a score $\scoreu_k$ to each response $\userc_k$ in the pool. Based on these scores, we define the distribution $p(\user|\context, \instr)$ as follows:
\begin{equation}
    p(\user|\context, \instr) = \begin{cases}
    \frac{\exp(\scoreu_k)}{\sum_{j=1}^r\exp(\scoreu_j)}, & \text{if } \user=\userc_k \in \poolu \\
    0, & \text{if } \user \notin \poolu
    \end{cases}
\end{equation}
The candidate response with the highest probability is selected as the next user utterance and sent to the agent bot. Next, we discuss the various modules in the user bot and how they are trained.

\subsubsection{Response Generator}
 The aim of response generator module is to output a pool of candidate user utterances for the given dialog history and the instructions.
 To achieve this, an autoregressive distribution over the tokens of the utterance $\user$ is defined. We finetune the pretrained GPT-2 network to model this distribution.
Specifically, given the tokens in the instructions and the dialog history, the GPT-2 network is trained to output the tokens of the user utterance. The utterance generated are in delexicalised format which are lexicalised from the values present in instruction before being shown to the agent bot(once selected by the selector) as shown in Figure \ref{fig:input_output_format}. 

While it is possible to sample an utterance from the GPT-2 network via greedy sampling or beam search, this poses several issues.
Firstly, autoregressive distributions tend to assign high probability to short utterances. Secondly, commonly occurring utterances in the corpus tend to have higher probability than the informative responses that are less frequent. We noticed that in lower data settings, the greedy response may not always be a relevant response. Nucleus sampling generates diverse responses which helps the response selector to pick more informative responses w.r.t the given context.

Hence, once the network has been trained, we sample multiple user responses from the network via nucleus sampling~\cite{nucleus_sampling} to obtain a pool of candidate responses $\poolu = (\userc_1, \ldots, \userc_r)$. This pool of candidates is fed to the response selector module as shown in Figure~\ref{fig:user_and_agent_bots}.

\subsubsection{Response Selector}
The aim of the response selector module is to assign a score to each candidate response in the pool based on its relevance to the context. We achieve this by feeding the tokens of the context and the candidate response(concatenating them with [CAN] token) to a Longformer network architecture~\cite{beltagy2020longformer}. 
The network outputs a contextualized embedding for each token. We feed the embedding of the [CLS] token through a linear layer followed by a sigmoid unit. The output of the network corresponds to the score assigned to the response for the given context. 

The network is trained to assign high scores to the positive (or ground-truth) responses while assigning low score to the negatively sampled responses. For each gold context-response pair, we provide a total of 10 negative response samples. These samples contain 5 random responses, 2 responses which are already part of the context (to stop the response selector from picking such responses) and 3 responses formed by concatenating 2 random responses to discourage the selector from picking longer candidate responses.

The network is trained via the triplet loss~\cite{image_similarity_loss,triplet_loss}. Specifically, given the context $\context$, the ground-truth response $\user_p$ and a negatively sampled response $\user_n$, the triplet loss is defined as follows:
\begin{equation}
    L(\context, \user_p, \user_n) = \max(0, \score(\context, \user_n) - \score(\context, \user_p) + \alpha)\,,
\end{equation}
where $\score(\context, \user)$ is the score assigned by the network to the response $\user$ for the given context $\context$. We use $\alpha=.05$ in our experiments. 

\subsection{Agent Bot}
The agent bot (distinct from user bot) models the distribution of the agent response $\agent$ conditioned on the context $\context$, the user utterance $\user$ and the knowledge base $\kb$, that is, $p(\agent| \context, \user, \kb)$. This distribution is modeled in four steps as shown in Figure~\ref{fig:user_and_agent_bots}.
Firstly, the agent bot feeds the context and the last user utterance to the {\em belief state generator} module which outputs a belief state of slot-value pairs (also referred to as query). Next, the query is executed over the knowledge base and a set of entities $\retrieved$, whose attributes match the values in the query, are returned. The total number of entities, the belief state, the dialog history and the previous user utterance are fed to the {\em response generator} which outputs a pool $\poola=(\agentc_1, \ldots, \agentc_r)$ of candidate responses in delexicalised format as seen in Figure \ref{fig:input_output_format}.  Finally, the responses in the pool are scored by the {\em response selector}. Based on these scores, we define the distribution of the agent response as follows:
\begin{equation}
    p(\agent|\context, \user, \kb) = \begin{cases}
    \frac{\exp(\scorea_k)}{\sum_{j=1}^m\exp(\scorea_j)}, & \text{if } \agent = \agentc_k \in \poola \\
    0, & \text{if } \agent \notin \poola
    \end{cases}
\end{equation}
where $\scorea_k$ is the score of the $k^{th}$ candidate response. Note that in equation, we do not show agent utterance being dependant on the belief state since it is calculated internally using context $\context$ and previous user utterance $\user$.
The candidate response with the highest probability is selected and sent to the user bot to generate the next turn. This interaction between the user and agent bots is repeated until the user bot outputs the end-of-dialogue token.

Next, we discuss in detail about the modules in the agent bot and how these modules are trained. Note that these modules do not share weights with the corresponding modules of the user bot.  

\subsubsection{Belief State (query) Generator}
The aim of the belief state generator is to generate a belief state for the given dialog history and last user utterance. Here, belief state contains the current domain followed by a sequence of key-value pairs of the form {\em $<$attribute\_name=attribute\_value$>$}.
To achieve this, we define a distribution 
over the belief states that can be executed over the knowledge base. 
The belief state generator treats the belief state as a sequence of tokens $\query = (\tokenq_1, \ldots, \tokenq_t)$. We train a GPT-2 network to model the distribution of the belief state tokens given the tokens of the dialog history and user utterance. 
Once the belief state generator has been trained, a belief state is sampled by greedy sampling and executed over the knowledge base.

\begin{figure}
\scriptsize
\fontsize{7.3pt}{10pt}\selectfont
\noindent\fbox{
    \parbox{.935\linewidth}{
        \textbf{Instruction} : You are looking for a  \textbf{particular hotel} .
        Its name is called  \textbf{bridge guest house}. Make sure you get  \textbf{hotel type}  and  \textbf{phone number} . \\
        
        \textbf{User:} \textcolor{darkgreen}{hi, i am looking for information on the \textbf{bridge guest house}.}
    }
}
\noindent\fbox{%
    \parbox{.95\linewidth}{%
        GREEDY RESPONSE: \\
        \textbf{Agent Response:} \textcolor{purple}{i have [value\_count] guesthouses and [value\_count] hotel -s that fit that criteria . do you have a preference for price range?}
    }%
}
\noindent\fbox{%
    \parbox{.95\linewidth}{%
        NUCLEUS DECODING SAMPLES(Top 5): \\
        \textbf{Response 1:} i have [value\_count] guesthouses available, how many are in your area or price range? \\
        \textbf{Score :} 0.54 \\
        \textbf{Response 2:} i have [value\_count] options, [hotel\_name] and [hotel\_name], both of which offer free wifi and parking. do you have any other preferences? \\
        \textbf{Score :} 0.31 \\
        \textbf{Response 3:} there are several guesthouses in the [value\_pricerange] price range. do you have a preference? \\
        \textbf{Score :} 0.54 \\
        \textbf{Response 4:} i have [value\_count] results. what area would you like to stay in? \\
        \textbf{Score :} 0.52 \\
        \textbf{Response 5:} \textcolor{darkblue}{i found the [hotel\_name], which is located on the [value\_area] side of town in the [value\_pricerange] price range. would you like to book a room?} \\
        \textbf{Score :} 0.89 
    }%
}

\caption{A goal along with context from the MultiWOZ dialog corpus where responses are generated using Greedy and Nucleus Sampling methods. Response highlighted in blue(highest score) was finally chosen by the model.
}
\label{fig:nucleus_greedy}
\end{figure}

\subsubsection{Response Generator}
This module mimics the response generator of the user bot with the exception that the input to the GPT-2 network comprises the context, the last user utterance, the belief state and the total number of KB entities satisfying the belief state. We provide only the number of entities instead of entire entities to the agent response. This is done as the response would be different for $0$, $1$ and more than $1$ matched entities and further information about entities could be filled while lexicalising the response. The GPT-2 network is used to define an autoregressive distribution over the tokens of the agent response and is trained using maximum likelihood. Once the module is trained, a pool of candidate responses $\poola$ is sampled via nucleus sampling. The response is lexicalised using the values from the belief state before being shown to the user bot. Figure~\ref{fig:nucleus_greedy} illustrates the advantages of using nucleus sampling for our decoders followed by use of a response selector.

\subsubsection{Response Selector}
This module outputs the score of each agent response in the candidate pool. To achieve this, the context, the last user utterance and the agent response are fed to the Longformer network architecture. The training of this network as well as the selection of negative samples mimics the training of the response selector for the user bot. Once the model has been trained, it outputs a score $\scorea$ for each agent response in the candidate pool. 

All the user and agent utterances, belief states and KB results created form the generated dialog.

\section{Experiments}
\label{experiments}

\subsection{Datasets}
To demonstrate the strength of our work we experiment on two different types of tasks - (i) Task oriented dialogs using the MultiWOZ 2.0 dataset \cite{Multiwoz}  (ii) Persona-based conversation generation using the PersonaChat dataset \cite{personachat}. 
\subsubsection{Task-Oriented Dialog} 
MultiWOZ \cite{Multiwoz} is a large scale multi-domain dialogue dataset consisting of 8438 training, 1000 validation and 1000 test conversations distributed across $7$ domains. Each conversation is associated with {\em instructions} which were were used by the crowd workers to generate the conversations. 
30\% of the dataset consists of conversations with a {\em single} goal while the rest are {\em multi-goal} dialogues, i.e, conversations accomplish more than one task -- example, booking a train followed by making a restaurant reservation.

\subsubsection{Persona-based Conversation} 
PersonaChat \cite{Persona} is a large scale non task-oriented dataset containing a set of 1155 distinct characters, each consisting of at least 5 profile sentences. The dataset is collected via Amazon Mechanical Turk where each of the pair of speakers condition their dialogue on a given profile, which is provided. It contains a total of 10,907 dialogs out of which 1000 dialogs are used for validation while 968 dialogs are used for testing.

\begin{table*}[h!]
\centering
\scriptsize
\begin{tabular}{|l|c|c|c|c|c|c|c|c|l|l|l|l|}
\hline
\multirow{2}{*}{ \hspace{30pt} MODELS} &
\multicolumn{4}{c|}{5\%}    & \multicolumn{4}{c|}{20\%}  & \multicolumn{4}{c|}{100\%}                                                               \\ \cline{2-13} 
                      & B     & I    & S    & C     & B     & I    & S    & C    & \multicolumn{1}{c|}{B} & \multicolumn{1}{c|}{I} & \multicolumn{1}{c|}{S} & \multicolumn{1}{c|}{C} \\ \hline
DAMD \cite{DAMD}       & 9.5   & 48.4 & 25.8 & 46.6  & 12.4  & 54.1 & 32.3 & 55.6 & 16.9                   & 72.7                    & 60.3                 &   83.5                 \\ \hline                        
DAMD-MADA \cite{DAMD}       & 9.5   & 50.5 & 33.9 & 51.8  & 13.1  & 60.1 & 42.9 & 64.6 & 16.6                  & 76.3                   & 60.4                   & 85.0                   \\ \hline 
PARG - TSCP~\cite{PARG}               & 13.1 & 53.5  & 39.2 & 59.4 & 13.0 & 63.6 & 48.9 & 69.2 & 15.4                   & 80.1                   & 63.1                   & 87.0                 \\ \hline\hline

Soloist \cite{Soloist}               & 12.5  & 52.5 & 32.9 & 55.2  & 14.0  & 60.9 & 50.0 & 69.5 & 16.8                   & 80.5                   & 63.2                  & 88.6                  \\ \hline
 Soloist (Paraphrase)             & 10.9   & \textbf{57.8} & 38.4  & 59.0 & 13.9  & 62.9 & 52.7 & 71.7 & 16.4                  &  \textbf{82.2}                 & 62.6                   &  88.8                \\ \hline 
\rowcolor{LightCyan} Soloist (Sim. Aug.)             & \textbf{14.0}  & 55.3 & \textbf{41.4} & \textbf{62.3}  & 14.8  & \textbf{70.5} & \textbf{56.4} & \textbf{78.3} & 17.6                   & 76.5                   & 60.9                   & 86.3                   \\ \hline \hline
MinTL-T5-Small \cite{MinTL}         & 12.5  & 50.9 & 33.9 & 55.7  & 15.8  & 63.5 & 48.8 & 72.0 & 17.4                   & 80.1                   & \textbf{64.7}                   & \textbf{89.8}                   \\ \hline
\rowcolor{LightCyan} MinTL-T5-Small (Sim. Aug)      & 13.1  & 57.6 & 36.1 & 60.0  & \textbf{16.0}  & 68.0 & 55.1 & 76.6 & \textbf{18.5}                   & 79.5                   & 57.1                   & 86.8                   \\ \hline 

\end{tabular}
\caption{Performance of models using varying sizes of MultiWOZ 2.0 dataset (B,I,S,C stand for BLEU, Inform, Success and Combined scores respectively). `Sim. Aug.' refers to the use of our simulated data.  Bold values indicate the highest scores.}.
\label{tab:models-data-2.0}
\end{table*}

\begin{table*}[h!]
\centering
\scriptsize
\begin{tabular}{|l|c|c|c|c|c|c|c|c|}
\hline
\multirow{2}{*}{\hspace{30pt} Model} & \multicolumn{2}{c|}{5\%} & \multicolumn{2}{c|}{10\%} & \multicolumn{2}{c|}{20\%} & \multicolumn{2}{c|}{100\%} \\ \cline{2-9} 
                      & PPL        & Hits@1      & PPL         & Hits@1      & PPL         & Hits@1      & PPL         & Hits@1       \\ \hline
GPT2-small             & 40.4      & 11.1        & 41.2       & 13.3        & 33.6       & 13.8        & 35.8        & 14.6         \\ \hline
\rowcolor{LightCyan} GPT2-small (Sim. Aug.)        & 41.1      & 12.2        & 39.1       & 14.5        & 40.2       & 14.0        & 43.0       & 15.9         \\ \hline
Lost In Conversation \cite{Lostinconversation}        &      -      &    -         &     -        &     -        &   -          &   -          &    -         &   17.3           \\ \hline
\end{tabular}
\caption{Performance of models using varying sizes of Persona dataset. PPL stands for Perplexity}.
\label{tab:persona}
\end{table*}

\subsection{Data Generation using Simulators}
\noindent {\bf MultiWoz: }As mentioned previously, our simulator allows the generation of new conversations based on {\em instructions}.
In our experiments, we operate our simulators using 5\% (421/8438), 20\% (1684/8438) and 100\% of the original training data. For 5\% and 20\%, we use the instructions of the remaining training datasets (i.e. remaining 95\% and 80\% respectively) to generate simulated conversations. The simulated conversations are added to the original conversations, thereby ensuring that the size of the datasets remains unchanged. In case of 100\% we train our simulators on the entire training data and then generate simulated conversations using the instructions of the same data. The simulated conversations are then appended to the original conversations. The resulting dataset has twice as many conversations as the original dataset.


Recall that each conversation requires KB and belief state by the agent. Our agent simulator generates queries for the simulated data using belief state generators as described earlier. While training the end-task dialog models using the simulated data, we use these generated values as the oracle belief state for our simulated data. Similar to existing work on this dataset, we use delexicalised agent utterances using the format followed by MultiWOZ \cite{Multiwoz} which are later updated with KB values based on the results of the query. Hyper-parameter settings are available in supplementary notes. 

\noindent{\bf PersonaChat: } In case of PersonaChat dataset, we train a single user bot to mimic both the users of a conversation. To generate the utterance for a specific user, the corresponding persona is fed to the bot along with the dialog context. Thus, a single bot is able to simulate a conversation between two distinct personas. We use 5\%(447/8939), 20\%(1788/8939) and 100\% of the training data in our experiments just like in MultiWoz. 


\subsection{End-Task Models}
\noindent{\bf MultiWoz: } We experiment with two recent end-task models: Soloist (initialized with GPT2-small) \cite{Soloist} and MinTL-T5 (initialized with T5-small) \cite{MinTL}. Soloist is a transformer based auto-regressive model that incorporates dialog modules, including the query generator, into a single network. The original model was pre-trained on a variety of dialog tasks and then applied to MutliWoz in few-shot settings. However, we use an untrained instance of Soloist, initialized only using GPT2-small, as our goal is only to demonstrate that simulated-data based augmentation can help train useful end-task models. MinTL-T5 is another recent model that also uses pre-trained transformers along with an improved method for updating belief states.

\noindent{\bf PersonaChat: } We use GPT2-small based end-task model to test effectiveness of simulated chat. 

\subsection{Baselines}
\noindent{\bf MultiWoz: } As baselines, we study the performance of our end-task models based on Soloist and MinTL-T5, when they are trained in the absence of data augmentation.  
We look at non-augmentation based recent baseline model DAMD \cite{DAMD}. Additionally, we compare the performance of our simulation based augmentation, against a recent approach - DAMD-MADA \cite{DAMD} which uses dialog-states based augmentation and PARG-TSCP \cite{PARG} which uses paraphrases to help improve response generation done by TSCP \cite{TSCP}. Additionally, we experiment with a T5 \cite{2020t5} based paraphrase generation model\footnote{https://huggingface.co/Vamsi/T5\_Paraphrase\_Paws} fine-tuned on the PAWS dataset \cite{paws2019naacl} -- we use this  model to generate paraphrases and augment training data and refer this model as Soloist(Paraphrase) in Table \ref{tab:models-data-2.0}(Details in Supplementary).

\noindent{\bf PersonaChat: } We compare the performance of an end-task model based on GPT2-small with and without augmented data. We report the performance of the `Lost in Conversation' model \cite{Lostinconversation}, the winner of the ConvAI2 challenge.\footnote{https://parl.ai/projects/convai2/} 

\subsection{Metrics}
{\bf MultiWoz: }We evaluate the usefulness of our generated data by using it to train a dialog model for the end-task. We therefore use BLEU (B), {$Inform$} (I) and {$Success$} (S) rates as defined by \citeauthor{Multiwoz}, along with Combined(C) score \cite{Shikib} given by, $BLEU + 0.5 \times (Inform + Success)$. While BLEU evaluates the fluency of the generated response, $Inform$ and $Success$ measure the relevance of the agent utterances. Specifically, the $Inform$ Rate measures the correctness of the entity provided by the agent, while the $Success$ Rate measures how often the agent was able to provide correct attributes when requested by the user. We note that there are minor (but significant) differences in delexicalization used by different models and this makes the direct comparison using the metrics inaccurate. In our experiments, we use the delexicalisation scheme used by \cite{Multiwoz} and their\footnote{https://github.com/budzianowski/multiwoz} task-evaluation scripts to report results. Hence we see slight drop in the scores in table \ref{tab:models-data-2.0} for Soloist and MinTL models compared to the scores cited in their respective papers(see suplementary for details). 

{\noindent \bf PersonaChat:} We use $Hits@1$ and $Perplexity$ as in \cite{Persona} to evaluate the models. While $Perplexity(PPL)$ measures the log likelihood of the correct sequence, $Hits@1$ scores the responses in a next-utterance (response) prediction task -- given an input context and persona, the goal is to predict the correct(ground-truth) response from a set consisting of other incorrect responses.

\subsection{Results}
We study the following research questions:  
(1) Would the new student model trained on simulated conversations along with crowd generated low(95\% + 5\%) and medium (20\% + 80\%) data perform better than teacher models trained only on only low(5\%) and medium(20\%) resourced data? (2) How does simulated-data based augmentation compare with recent work on augmentation? (3) How does the student model perform compared to models trained on 100\% human generated data? (4) Can we use this technique to improve the models trained already on 100\% human generated data?  (5) What is the qualitative difference between simulated and crowd-sourced chats? 


\subsubsection{Use of simulated data in End-task} 

\noindent{\bf MultiWoz: } 
Table \ref{tab:models-data-2.0} shows the use of simulated data helps improve performance in low data settings (5\%) and medium data settings (20\%). The use of simulated data helps improve performance of both Soloist and MinTL-T5  (gains of 7-8\% in combined metric) in 5\% data setting. We also see a higher improvement in medium data setting i.e. an increase of 7\% in combined score for MinTL  and 13\% for Soloist model suggesting the effectiveness of our method in low and medium resource setting. 

We further compare the performance of our augmented data w.r.t the original 100\% dataset for both Soloist and MinTL. Adding additional 80\% augmented data to our Soloist model trained on 20\% dataset substantially increases the combined score from 69.5 to 78.3 although it lags behind the model when trained on 100\% human generated dataset which gives a combined score of 88.6. Similarly for MinTL, the combined score improves from 72 to 76.6 but falls short of the performance on original 100\% human generated dataset.  Our model behaves according to the common knowledge that noisy student models do not perform as good as a teacher model trained on a similar sized data(100\%). Through this experiment we show how the diverse knowledge contained in Longformer could be transferred to our augmented dataset by using it as a selector. The teacher generator model generates a list of diverse candidate responses(by top-p sampling) which is provided to the selector to pick the most relevant response helping the selector induce its knowledge through the process. Thus we see that the method improves the models trained on low-medium sized datasets  and can be used effectively when larger datasets are not available.

Our data generation technique is not able to increase the performance of models trained on 100\% human generated data( i.e.200\% not performing better than 100\%) the reasons for which are discussed under human study section. Additional qualitative results on the MultiWOZ dataset are available in the supplementary material. 

Comparing with other baseline models, student models trained on simulated data(Sim. Aug.) on Soloist and MinTL outperforms existing end-task models such as dialog-state based augmentation (DAMD-MADA) and paraphrase-based augmentation(PARG and Soloist(paraphrase)) in low and medium data settings as seen in Table \ref{tab:models-data-2.0}. Soloist (Sim. Aug) gets combined score of 62.3 in 5\% data compared to 59.4 obtained by the best performing augmentation based baseline model PARG-TSCP. Similarly Soloist (Sim. Aug) scores 78.3 in 20\% data compared to 69.2 obtained by PARG-TSCP.

\noindent{\bf Persona Chat: } Table \ref{tab:persona} shows improvement in \textit{Hits@1} when the GPT2-small based end-task model is trained on simulation-based augmented data. Gain in \textit{Hits@1} (2-10\%) demonstrates that the model is able to learn the context and persona of the given characters better which results in better conversations. The augmented data helps improve the performance of a simple GPT2-small model(fine-tuned on dataset) in \textit{Hits@1} from 14.6 to 15.9 which is very close to 17.3 achieved by Lost in Conversation. The Perplexity (PPL) gives mixed results suggesting that the language {\em style} of simulated conversations differs from  the language {\em style} of the original dataset. This is because GPT2-small incorporates its pre-trained knowledge in the simulated conversations. However the fact that \textit{Hits@1} consistently increases across all dataset sizes suggests that the generated simulated conversations help the model capture the context and persona better despite changing its language {\em style}.  

\begin{table}[h!]
\centering
\scriptsize
\begin{tabular}{|l|c|c|}
\hline
                  & \multicolumn{1}{l|}{Original Data} & \multicolumn{1}{l|}{Simulated Data} \\ \hline
Relevance         & 4.7                                & 4.0                                 \\ \hline
Grammar           & 4.6                                & 4.5                                 \\ \hline
User Bot Fluency  & 4.5                                & 4.1                                 \\ \hline
Agent Bot Fluency & 4.6                                & 4.1                                 \\ \hline
\end{tabular}
\caption{Human evaluation scores(scale of 1-5) on original and simulated data}.
\label{tab:human_eval}
\end{table}

\subsubsection{Human Study} 
To assess the qualitative difference between simulated data and crowd-sourced data, we conducted a blind human-study involving six participants. Participants were presented the crowd-annotation instructions from MultiWOZ and were asked to assess the quality of a pair of dialogs corresponding to the same instruction -- one generated by the crowd workers (from the original dataset) and the other generated by our simulators. 
The participants were blind to the source of the dialog (crowd or simulator). Each dialog was scored on the Likert scale(1-5) by answering the following questions: 1) `{How relevant is the dialog w.r.t the dialog generation instruction?}' 2) `{How good is the grammar of the utterances?}' 3) `Are the {\em user} utterances responding to {\em agent} utterances fluently and meaningfully?' 4) `Are the {\em agent} utterances responding to {\em user} utterances fluently and meaningfully?'.  Each participant evaluated $25$ pairs resulting in a total evaluation set of $150$ pairs. 

As seen in Table~\ref{tab:human_eval}, the simulated data is of high quality with the bots scoring well on fluency as well as grammar.  As expected, there is a slight deterioration in relevance to the instructions compared to crowd-sourced conversations. This happens because the simulated conversations may not use all the information present in the instructions. This also answers why the simulated data doesn't increase the performance of models trained on 100\% dataset in Table \ref{tab:models-data-2.0}. In lower data settings, the original dialogs of the remaining instructions (i.e. remaining 95\% in case of 5\% training data) were not part of dialog used for the end-task model. Hence, the simulated data provided new dialogs that  were never seen by the model. In 100\% data setting,  since the model  had already seen the original dialogs, the simulated  dialogs did not improve the performance as they lacked some relevance w.r.t the instructions when compared with human generated data. The same issue causes the model trained on 100\% original dataset  to perform better than our augmented datasets i.e. (20+80)\% and (5+95)\%.

\section{Conclusion}
In this paper, we demonstrated a dialog generation framework that mimics the data creation process employed by crowd workers. We find that our method is able to generate meaningful conversations that aids the training of end-task dialog models in low resource data settings. The use of additional simulated data to train end-task dialog models result in a performance improvement of 7-13\% in low resource settings of MultiWOZ 2.0 dataset and 2-10\% increase in Hits@1 in case of PersonaChat. 
The simulation-framework does not make strict assumptions about the domain or dataset and can be applied to diverse dialog tasks such as task-oriented dialog and persona-based chat. In future, it would be interesting to compare the strengths of different augmentation methods and how they may be effectively combined. 

\bibliography{emnlp2021}
\bibliographystyle{emnlp2021}

\appendix

\section{Appendix Overview}
Section \ref{Hyperparameter} provides information on the hyperparameter settings of models used in the experiments. Section \ref{cost} does the cost analysis and Section \ref{eval} discusses about the inconsistencies in evaluation of various models in details. The paraphrase based augmentation method used to train an end-task Soloist \cite{Soloist} model has been described in Section \ref{para}.  We also show further experimental results in Section \ref{2.1} and \ref{exp} while additional qualitative study is shown in Section \ref{qual}.

\section{Hyperparameter Settings} \label{Hyperparameter}
\noindent {\bf MultiWoz: }We create separate user bots and agent bots along with their constituent modules consisting of query models (for tracking belief state), response generators and response selectors.
We use GPT2-small (12 layered, 768 hidden size, 117M parameters) from the `Transformers' library by Huggingface \cite{Wolf2019HuggingFacesTS} for the response generator . For response selectors, we use Longformers (12 layered, 1024 hidden size, 149M parameters) \cite{beltagy2020longformer} for both user and agent models. We train on 5\%, 10\%, 30\% and 100\% of the training data with a learning rate of 1e-5. Adam optimizer with default settings is used for all the models. 

\noindent {\bf PersonaChat: }Similar to MultiWOZ, the response generators use GPT2-small while response selectors use Longformers. There is no belief state generator and only single user model is trained i.e. no separate agent model exists. Adam optimizer is used with a learning rate of 1e-5.  

\section{Cost comparison} \label{cost}

The response generator, belief state generator and response selector models(total 5) each take 1 day on a single V100 GPU to generate the dialogues(for 100\% data). MultiWOZ data creation, on the other hand, required 1249 workers for the entire process. An Amazon EC2 P3 instance costs \$3.06 per hour in an On-Demand setup costing less than \$400 for the entire process. Generating 10.4K dialogs with 1249 workers (2 workers per conversation) means 15-16 dialogues per worker and assuming they take an hour to generate the conversation with a minimum wage of \$6 per hour payment, it leads upto \$7.5k. Our method is clearly both cost and time effective when compared with the crowdsource workers.

\section{Evaluation Inconsistency} \label{eval}
We noticed the delexicalisation used in models such as PARG \cite{PARG}, DAMD \cite{DAMD} and MinTL \cite{MinTL} was different from the delexicalisation used in original MultiWOZ code. Since, the delexicalised agent responses are used by the official evaluation script to score the dialogs there were discrepancies in the evaluation of different models. Some delexicalisation differences included use of tokens such as [restaurant\_phone] or [hotel\_address] in original format while some models used [value\_phone] for such requestables. This lead to difference in evaluation script since the official script looks at the domain of requestable i.e. [train\_address] should have been used if the utterance domain is 'Train' and not [hotel\_address]. Replacing these tokens with a generic [value\_address] as used in other models reduces the complexity of such dialogs leading to higher innform and success scores. The Soloist and MinTL models used in our experiments are  trained on the delexicalisation used in the MultiWoz original code\footnote{https://github.com/budzianowski/multiwoz}.  This leads to a small drop in the overall performance of the models.

\section{Soloist Paraphrase Model} \label{para}

In order to train a soloist \cite{Soloist} end task model using paraphrases generated from original training data, we use T5 \cite{2020t5} based paraphrase generation model\footnote{https://huggingface.co/Vamsi/T5\_Paraphrase\_Paws} fine-tuned on the PAWS dataset \cite{paws2019naacl}. We also add training paraphrase data from PARG \cite{PARG} to create a mixture of corresponding paraphrases for each utterance. In low resource setting(5\% and 20\% of training dataset) we generate enough paraphrases to take the whole size of augmented data to 100\% of original training data. i.e. We add paraphrases equaling the size of 95\% of total data in case of 5\% and 80\% of total data in case of 20\%. Adding paraphrases from T5 fine-tuned model and PARG provides the model with diversity inn training data for final end-task model. 

\section{MultiWoz 2.1 Performance} \label{2.1}
In order to check our performance on the revised version of MultiWoz, we experimented with MultiWoz 2.1 as well. Since not many baseline models have results on this version of the dataset, we used MultiWOZ 2.0 as the dataset to compare our model with other baselines.  Table \ref{tab:main-results} presents results using the Soloist end-task model for MultiWOZ 2.1. Additional data generated by our simulators results in a significant improvement on the Combined metric for both the {\em oracle} belief states(O) and the {\em generated} belief states(G) in the end-task model. The use of our simulated data results in a 18-25\% improvement in combined metrics for low-medium resource setting. The performance of generated belief states shows that the query generator produces accurate queries most of the time.

\begin{table*}[h!]
\centering
\scriptsize
\begin{tabular}{|c|c|c|c|c|c|c|c|c|c|}
\hline
\multirow{2}{*}{Datasize} & \multirow{2}{*}{Beliefstate} &
\multicolumn{4}{c|}{Without Augmentation} & \multicolumn{4}{c|}{With Augmentation} \\ \cline{3-10} 
                                                                        &                                                                            & BLEU (B)        & Inform (I)      & Success (S)      & Comb. (C)       & BLEU (B)       & Inform (I)      & Success (S)     & Comb. (C)      \\ \hline
5\%                                                                     
& O
& 12.5     & 47.3   & 33.8   & 53.1    & 12.2   & 67.8   & 40.4  & \textbf{66.3}  \\ \hline
10\%                                                                     
& O                                                                     & 13.8     & 51.9   & 37.6   & 58.5    & 13.8   & 68.1   & 49.5  & \textbf{72.6}  \\ \hline
20\%                                                                     
& O                                                                   & 15.2     & 61.3   & 48.7   & 70.2    & 13.7   & 81.8   & 64.1  & \textbf{86.7}  \\ \hline
100\%                                                                   
& O                                                                    & 16.4     & 85.8   & 74.4   & \textbf{96.5}    & 17.2   & 80.1   & 67.7  & 91.2  \\ \hline \hline
5\%                                                                     
&G
& 12.5    & 39.6   & 26.8   & 45.6   & 12.0   & 54.8   & 32.3  & \textbf{55.6}  \\ \hline
10\%                                                                      
&G                        & 13.5    & 46.2   & 31.9   & 52.6   & 13.7   & 52.6   & 38.5  & \textbf{59.3}  \\ \hline
20\%                                                                      
&G                        & 15.1    & 55.4   & 42.8   & 64.2   & 13.5   & 71.2   & 54.7  & \textbf{76.5}  \\ \hline
100\%                                                                     
&G                        & 15.9    & 80.1   & 64.3   & \textbf{88.1}   & 16.2   & 75.8   & 58.5  & 83.4  \\ \hline
\end{tabular}
\caption{Performance of the query generator and the end-task model Soloist \cite{Soloist} when trained on different dataset sizes with and without the use of additional simulated data for MultiWOZ 2.1. `O' and `G' in column two refer to the use of `oracle' and  `generated' belief states respectively.}
\label{tab:main-results}
\end{table*}

\begin{figure}[h!] 
\small
\fontsize{7.3pt}{10pt}\selectfont
\noindent\fbox{
    \parbox{.984\linewidth}{
        \textbf{ORIGINAL INSTRUCTIONS} : You are looking for a  \textbf{restaurant} . The restaurant should be in the  \textbf{expensive}  price range and should serve  \textbf{italian}  food . Once you find the  \textbf{restaurant}  you want to book a table for  \textbf{5 people}  at  \textbf{11:30}  on  \textbf{sunday}  . If the booking fails how about  \textbf{10:30}  . Make sure you get the  \textbf{reference number}  .\\
        
        \textbf{User:} \textcolor{darkgreen}{hi, i'm trying to find an \textbf{expensive italian} restaurant.} \\
        \textbf{Agent:} \textcolor{darkblue}{i have found several [value\_pricerange]  [value\_food] restaurant -s . is there a specific area of town that you are looking for ?}\\
        \textbf{User:} \textcolor{darkgreen}{i would like to go to the \textbf{south}, please.} \\
        \textbf{Agent:} \textcolor{darkblue}{there is an [value\_pricerange] restaurant called [restaurant\_name] in the [value\_area] . would you like to make a reservation?}\\
        \textbf{User:} \textcolor{darkgreen}{yes, i'd like to book a table for \textbf{5} at \textbf{11:30} on \textbf{sunday}.} \\
        \textbf{Agent:} \textcolor{darkblue}{i am sorry , but we did not get that reservation . would you like to try another day or time slot?} \\
        \textbf{User:} \textcolor{darkgreen}{can you try for \textbf{10:30} instead?} \\
        \textbf{Agent:} \textcolor{darkblue}{booked at [restaurant\_name] for [value\_count] people . reference number is [restaurant\_reference] .}\\
        \textbf{User:} \textcolor{darkgreen}{thank you so much.  that's all i needed.  bye.} \\
        \textbf{Agent:} \textcolor{darkblue}{thank you for contacting towninfo centers . enjoy your dining experience.}
    }
}
\noindent\fbox{%
    \parbox{\linewidth}{%
        \textbf{PERTURBED INSTRUCTIONS} : You are looking for a  \textbf{restaurant} . The restaurant should be in the  \textcolor{violet}{cheap}  price range and should serve  \textcolor{violet}{indian}  food . \textcolor{violet}{Restaurant should be in  \textbf{north}.}  Once you find the  restaurant  you want to book a table for  \textbf{5 people}  at  \textbf{11:30}  on  \textbf{sunday}. If the booking fails how about  10:30. Make sure you get the  \textbf{reference number}.
        
        \textbf{User:} \textcolor{darkgreen}{hi! can you help me find a \textbf{cheap indian} restaurant in the \textbf{north}?} \\
        \textbf{Agent:} \textcolor{darkblue}{sure ,  [restaurant\_name] is in the [value\_area] part of town . it is located at [restaurant\_address].}\\
        \textbf{User:} \textcolor{darkgreen}{i would like to book a table for \textbf{5 people} at \textbf{11:30} on \textbf{sunday}.} \\
         \textbf{Agent:} \textcolor{darkblue}{i am sorry booking was unsuccessful , do you need to find another restaurant ?}\\
        \textbf{User:} \textcolor{darkgreen}{can you try for \textbf{10:30} ?} \\
        \textbf{Agent:} \textcolor{darkblue}{i have booked you a table for [value\_count] people at [value\_time] on [value\_day] . your reference number is [restaurant\_reference] .}\\
        \textbf{User:} \textcolor{darkgreen}{thank you so much.  that is all i needed today.  bye.} \\
        \textbf{Agent:} \textcolor{darkblue}{enjoy your meal !} 
    }%
}

\caption{Examples of generated samples using original instructions from MultiWOZ dataset against perturbed instructions with minor changes. 
The generated dialogues show the robustness of our generator model which is able to generate an entirely new conversation with slight variations in the goal.
}
\label{fig:Perturbed_goal}
\end{figure}

\section{Single Goal Dialogs} \label{exp}

\begin{table*}[h!]
\centering
\begin{tabular}{|c|l|c|c|c|c|c|c|c|c|}
\hline
Dataset                & Belief State & \multicolumn{2}{c|}{BLEU} & \multicolumn{2}{c|}{Inform} & \multicolumn{2}{c|}{Success} & \multicolumn{2}{c|}{Combined} \\ \hline
\multicolumn{1}{|l|}{} &              & w.          & w/o.        & w.           & w/o.         & w.            & w/o.         & w.            & w/o.          \\ \hline
5\%                    &  Oracle   &  9.3           &   8.0          &  88.5            &    82.0          &       64.1        &    60.6          &   85.7            &  79.4             \\ \hline
10\%                   &  Oracle  & 10.8       & 10.8       & 92.5        & 84.5        & 75.2         & 69           & 94.7          & 87.6         \\ \hline
30\%                   & Oracle    & 11.9       & 12.4       & 90.3        & 82.3        & 73.0         & 65.5        & 93.6         & 86.3         \\ \hline
100\%                  & Oracle    & -           & 14.9       & -            & 82.8         & -             & 78.3        & -             & 95.5         \\ \hline
\end{tabular}
\caption{Score against single goal conversations in test dataset with oracle belief state}
\label{tab:1}
\end{table*}

\begin{table*}[h!]
\centering
\begin{tabular}{|c|c|c|c|c|c|c|c|c|c|}
\hline
Dataset                & \multicolumn{1}{l|}{Belief State} & \multicolumn{2}{c|}{BLEU} & \multicolumn{2}{c|}{Inform} & \multicolumn{2}{c|}{Success} & \multicolumn{2}{c|}{Combined} \\ \hline
\multicolumn{1}{|l|}{} & \multicolumn{1}{l|}{}             & w/o.          & w.        & w/o.           & w.         & w/o.            & w.         & w/o.            & w.          \\ \hline
5\%                    & Oracle                            & 7.1            &  9.2           &   63.2           &   73.2           &    34.4           & 42.6             &     55.9          &  67.1             \\ \hline
10\%                   & Oracle                            & 9.6       & 10.8        & 63.8         & 78.2         & 38.9          & 52.9         & 61.0         & 76.4         \\ \hline
30\%                   & Oracle                            & 9.8       & 12.4        & 66.6           & 77.0         & 34.9          & 52.3         & 60.6          & 77.1         \\ \hline
100\%                  & Oracle                            & 15.9           & -       & 72.8            & -         & 63.7             & -         & 84.2             & -          \\ \hline
\end{tabular}
\caption{Score against entire test dataset containing both single and multiple goal conversations with oracle belief state}
\label{tab:2}
\end{table*}

We also evaluate the performance of our model when trained on single goal dialogs of MultiWOZ 2.1 to test whether the model can learn generating multiple goal dialogs using just single goal data. We train separate models(generators and selectors) for each domains and the simulated single dialog chats were combined (pair of dialogs) from different domains using a basic script and trained the final end-task model. Our method achieves significant improvement over non-augmented(w/o.) dataset as seen in Table ~\ref{tab:1}, ~\ref{tab:2}, ~\ref{tab:3} and ~\ref{tab:4}. Tables ~\ref{tab:1} and ~\ref{tab:3} show the performance of the model for the oracle and generated belief state on single domain goal conversations of the test dataset. We see an improvement of 7-8 \% in combined score across all dataset sizes on applying our augmentation technique(w.). We are able to achieve a combined score of 94.7\% with just 10\% of the dataset which is very close to the combined score of 95.53\% when trained on the entire dataset. \\

\begin{table*}[h!]
 \centering
\begin{tabular}{|c|c|c|c|c|c|c|c|c|c|}
\hline
Dataset                & \multicolumn{1}{l|}{Belief State} & \multicolumn{2}{c|}{BLEU} & \multicolumn{2}{c|}{Inform} & \multicolumn{2}{c|}{Success} & \multicolumn{2}{c|}{Combined} \\ \hline
\multicolumn{1}{|l|}{} & \multicolumn{1}{l|}{}             & w/o.          & w.        & w/o.           & w.         & w/o.            & w.         & w/o.            & w.          \\ \hline
5\%                    & Generated                         & 7.6            &  9.6           &    40.0          &  80.9            &     30.1          &   60.6           &   42.7            &   80.4            \\ \hline
10\%                   & Generated                         & 10.7       & 10.7        & 77.4        & 84.5         & 61.5         & 66.3         & 80.1         & 86.2         \\ \hline
30\%                   & Generated                         & 12.2       & 11.8        & 77.8         & 84.1         & 60.2         & 63.3        & 81.2         & 85.5         \\ \hline
100\%                  & Generated                         & 14.8        & - & 81.4    & -    & 76.1  & -  & 93.5             & -         \\ \hline
\end{tabular}
\caption{Score against single goal conversations in test dataset with generated belief state}
\label{tab:3}
\end{table*}

\begin{table*}[h!]
\centering
\begin{tabular}{|c|c|c|c|c|c|c|c|c|c|}
\hline
Dataset                & \multicolumn{1}{l|}{Belief State} & \multicolumn{2}{c|}{BLEU} & \multicolumn{2}{c|}{Inform} & \multicolumn{2}{c|}{Success} & \multicolumn{2}{c|}{Combined} \\ \hline
\multicolumn{1}{|l|}{} & \multicolumn{1}{l|}{}             & w/o.          & w.        & w/o.           & w.         & w/o.            & w.         & w/o.            & w.          \\ \hline
5\%                    & Generated                         &  6.8           & 9.8            &  19.3            &    54.7          &    10.2           &   31.9           &   21.6            &    53.1           \\ \hline
10\%                   & Generated                         & 9.5       & 10.7        & 52.3         & 61.2         & 29.9          & 40.6         & 50.6         & 61.6         \\ \hline
30\%                   & Generated                         & 9.5       & 12.4        & 50.9         & 59.4         & 24.9          & 38.3         & 47.4         & 61.2         \\ \hline
100\%                  & Generated                         & 15.9           & -       & 66.2            & -         & 55.4             & -         & 76.7             & -         \\ \hline
\end{tabular}
\caption{Score against entire test dataset containing both single and multiple goal conversations with generated belief state}
\label{tab:4}
\end{table*}

Tables ~\ref{tab:2} and ~\ref{tab:4} show the performance of the model when we use the oracle and generated belief state on the entire test dataset. We see a massive improvement in both the oracle and generated belief state setting. While the oracle belief state results improve the combined score by 20.02\%, 25.2\% and 17.98\% for 5\%, 10\% and 30\% of the dataset respectively, we see an even bigger improvement of 146.11\%, 21.82\% and 29.12\% when using generated belief states for 5\%, 10\% and 30\% of dataset respectively. The augmentation helps in improving the combined score by a huge margin thus bringing them close to the score of entire dataset(100 percent). The results show that simulated data generated from single goal dialogs can also do a good job at generalising to multiple goal dialogs. This insight would be useful in combining various single goal dialogs from different datasets. 

\section{Qualitative Study - Instruction Perturbation} \label{qual}

We now present a qualitative study demonstrating how our simulator is able to accommodate changes to instructions and reflect them in a conversation. 
Figure ~\ref{fig:Perturbed_goal} shows the generated dialogs from an original instruction in MultiWOZ and another from instructions created by perturbing the original instructions. The generated dialogs demonstrate the robustness of our generator model which is able to produce new and meaningful conversations using new entities from the perturbed instructions. 
Further, the dialogues generated are very different from each other which shows the wide variety of conversations the simulators are capable of producing, when provided with similar goals. 

\end{document}